\documentclass[11pt,a4paper]{article}
\usepackage{arxiv}
\usepackage[english]{babel}
\usepackage[T1]{fontenc}
\usepackage[utf8]{inputenc}
\usepackage{graphicx}
\usepackage{amsfonts}
\usepackage{amsmath}
\usepackage{authblk}
\usepackage{subfigure}
\usepackage{caption}
\usepackage{icomma}
\usepackage[numbers]{natbib}
\usepackage{hyperref}
\usepackage[ruled,vlined]{algorithm2e}

\newcommand{\Mod}[1]{\ (\mathrm{mod}\ #1)}

\begin{document}
\title{Learning Numeracy: Binary Arithmetic with Neural Turing Machines}
\author[1 2]{Jacopo Castellini}
\affil[1]{At the time of work: Dept. of Mathematics and Computer Science, University of Perugia}
\affil[2]{Currently: Dept. of Computer Science, University of Liverpool}
\affil[ ]{\small J.Castellini@liverpool.ac.uk}
\date{}
\maketitle

\begin{abstract}
One of the main problems encountered so far with recurrent neural networks is that they struggle to retain long-time information dependencies in their recurrent connections. Neural Turing Machines (NTMs) attempt to mitigate this issue by providing the neural network with an external portion of memory, in which information can be stored and manipulated later on. The whole mechanism is differentiable end-to-end, allowing the network to learn how to utilise this long-term memory via stochastic gradient descent. This allows NTMs to infer simple algorithms directly from data sequences. Nonetheless, the model can be hard to train due to a large number of parameters and interacting components and little related work is present. In this work we use NTMs to learn and generalise two arithmetical tasks: binary addition and multiplication. These tasks are two fundamental algorithmic examples in computer science, and are a lot more challenging than the previously explored ones, with which we aim to shed some light on the real capabilities on this neural model.
\end{abstract}


\section{Introduction}
Computer programs are composed of three fundamental mechanisms: elementary operations, logical flow control and memory usage. In the history of neural networks \cite{history} only the use of elementary operations have been extensively explored since so far, but during the last few years the coupling with an external piece of memory is gaining popularity \cite{mnet}. Neural Turing Machines (NTMs) were developed in 2014 at Google DeepMind Labs \cite{ntm} in an attempt to couple a neural network with an external memory component in order to improve long-term dependency learning in sequences prediction. Although recurrent neural networks (RNNs) are Turing-complete on their own \cite{rnn}, the difficulties that arise during their training (like the vanishing or the exploding gradient problems \cite{problems,bptt}) prevented them from being employed in learning more complex tasks, for example algorithmic ones \cite{algorithms}.

NTMs derive their name from the analogy with standard Turing Machines (TMs) \cite{tm} in addressing an infinite (or at least large enough to be considered so) portion of memory with an attentional mechanism similar to the read/write head of a TM. In contrast to a standard TM, a NTM is a "differentiable computer" that can be trained using gradient descent methods and can therefore learn its own "program" independently (attempts using Neuroevolution \cite{evolution} and reinforcement learning \cite{rl} have also been made).

In human brains, the most similar process to an algorithm is the concept of "working memory" \cite{working}: this mechanism allows the brain to rapidly create "variables" \cite{variables} by storing short-term information and manipulating them in a rule-based way \cite{limitations}. The analogy with an algorithm is evident, and a NTM is similar to this process because it can learn tasks in which it is required to manipulate rapidly-created variables. Also the attention mechanism in a NTM is similar to the way the working memory bounds its information in certain slots of memory in the brain \cite{brain}, despite the fact that a NTM autonomously learns how to do that.

The purpose of this work is to investigate the capabilities of this new model, for which so little work has been done so far. We will focus on two basic but yet non-trivial algorithmic tasks: binary addition and binary multiplication. These tasks require a more complex and structured interaction with the memory than the previously considered tasks in order to be learned and generalized. The remainder of this work is organized as follows: Section \ref{related} briefly describes some related works on using neural networks to learn binary arithmetic, Section \ref{ntm} depicts and illustrate the model of a Neural Turing Machine, Section \ref{experiments} describes our experimental setting and presents results, and lastly Section \ref{conclusions} draws some conclusions and highlights possible future directions.

\section{Related Works}
\label{related}
Binary arithmetic has been a fundamental part of computer science since its very beginning. Being able to develop logical circuits to perform these operations for fixed-size numbers has driven many improvements in hardware development as well as computational boolean logic and circuits design. These circuits can then be further combined together to compute the value of arbitrarily large and complex functions, rendering them a fundamental component of automatic computation \cite{logic}.

Previous works investigated the task of learning binary arithmetic with neural networks. About the binary addition task, \cite{add} showed three depth-optimal feedforward neural networks able to perform $n$-bits sums. More recently, \cite{adder} developed a new model to perform addition in a more parallel way. Similarly, the multiplication task was studied in \cite{mul}, where a model optimal in depth was developed. Also \cite{multiplier} studied this task, developing a model that is more efficient and fast than the previous ones. \cite{arithmetic} studied both the problems using feedforward neural networks, founding solutions that are optimal in the depth of the network and bounds polynomially the number of neurons and synapses.

The common problem to all these models is that they have not any generalization capability: the networks are trained with $n$ bits long binary numbers and they only learn how to operate with $n$ bits, being not able to generalize what they have learned to larger numbers. Differently from these approaches, \cite{ngpu} presents a model based on different gated recurrent unit (GRU) layers and kernel operations that is able to generalize almost perfectly to larger sequences of bits than those it was trained on on both tasks, but the model is quite complex and requires many layers and an accurate parameters setting in order to work properly.

\section{Neural Turing Machines}
\label{ntm}
\begin{figure}[htbp]
\centering
\includegraphics[width=0.7\textwidth]{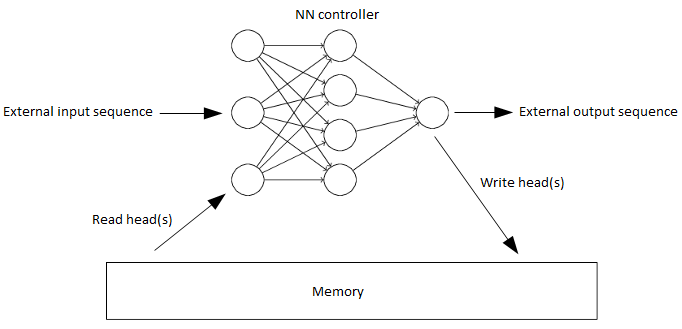}
\caption{The schematic structure of a NTM. The controller receives both an external input and some data read from the memory, process that and produce an output sequence, also eventually storing some data on the memory.}
\label{fig:ntm}
\end{figure}

Figure \ref{fig:ntm} shows the basic structure of a Neural Turing Machine. A neural network controller is trained to produce a desired output sequence from both an input sequence and the data stored in the memory. This controller can be either a feedforward or a recurrent network. By using an neural network as the controller, the NTM can learn how to manipulate the external input and memory to produce the correct output. This is in contrast with standard Turing Machines, which are created with a fixed and known program (its transition function). The external memory is a matrix of size $N\times M$ with $N$ locations of size $M$.

In order to be differentiable, the read and write operations are defined in a "blurry" way: instead of interacting with a single memory location (a row of the matrix) at each time step $t$, some selective read and write operations, called heads for analogy with TMs, interact to a different degree with every location of the memory. This degree is decided by an attentional focus mechanism that weights the heads via a normalized vector over memory rows, one for each read and write head. In this way, a head can look sharply at a single memory location or weakly over a number of locations.

\subsection{Memory Operations}
Let $M_t\in\mathbb{R}^{N\times M}$ be the content of the memory matrix at time step $t$, where $N$ and $M$ are the number of memory rows and the size of each row respectively, and let $w_t\in\mathbb{R}^N$ be the normalized weighting over the $N$ matrix rows produced by a read or write head at the same time step:

\begin{equation}
\sum_{i=1}^Nw_t(i)=1\qquad 0\leq w_t(i)\leq 1.
\end{equation}

Hence, the read vector $r_t\in\mathbb{R}^M$ produced by a read head at the time step $t$ is computed as:

\begin{equation}
r_t=\sum_{i=1}^Nw_t(i)M_t(i),
\end{equation}

where $M_t(i)$ is the $i$-th row of the matrix $M_t$. So, the read operation is differentiable with respect to both the weighting and the memory.

Inspired by the gate mechanism of a long-short term memory (LSTM) \cite{lstm}, the write operation is divided in two steps: an erase operation followed by an add operation. Let $e_t\in\mathbb{R}^M$ be the erase vector composed of $M$ elements in the range $(0,1)$ and $a_t\in\mathbb{R}^M$ the add vector produced by the write head, each matrix row $M_{t-1}(i)$ is modified as:

\begin{equation}
\tilde{M}_t(i)=M_{t-1}(i)\odot[\mathbf{1}-w_t(i)e_t],
\end{equation}
\begin{equation}
M_t(i)=\tilde{M}_t+w_t(i)a_t.
\end{equation}

where $\mathbf{1}$ is a vector of all $1$'s and $\odot$ is the element-wise product. So, a memory location is reset only if both the weight $w_t(i)$ and the erase vector $e_t$ are $1$, while it is left unchanged if at least one of them is $0$. In case of multiple write heads, the order in which the erase and the add operations are performed is irrelevant, because multiplication is commutative. The final value of the memory at time step $t$ is that obtained after all the write operations. Since both operations are differentiable, the combined operation is differentiable too. Because $e_t$ and $a_t$ are vectors of dimension $M$, these operations allow for a full control on which elements in each memory location have to be modified.

\subsection{Addressing Mechanisms}
The weightings $w_t$ are produced combining two different and complementary mechanisms: a content-based addressing that focuses the attention on locations with similar values with respect to the ones produced by the controller, and a location-based addressing in which data is retrieved based on its presence and location in the memory. The first addressing mechanism has the advantage of being simple, merely requiring the controller to emit an approximation of what had to be searched into the memory and then compare this approximation to retrieve the correct value. On the other hand, the location-based addressing mechanism allows to deal with tasks, such as arithmetical ones, for which the value of a variable is not important, but its presence in the memory and its location are. Content-based addressing is more general than the location-based one, because information about the data location can itself also be stored in the memory, but providing also the second mechanism as a primitive proved good for generalizing certain tasks, allowing easily for iterations through the locations and random access jumps.

In the content-based addressing, each read or write head first produces at time step $t$ a key vector $k_t\in\mathbb{R}^M$ and then compares it to every row $M_t(i)$ of the memory by using a similarity measure $K[\cdot,\cdot]$, like the cosine similarity, to produce a normalized weighting $w_t^c$ as:

\begin{equation}
w_t^c(i)=\frac{\exp\left(\beta_tK[k_t,M_t(i)]\right)}{\sum_{j=1}^N\exp\left(\beta_tK[k_t,M_t(j)]\right)},
\end{equation}

where $\beta_t$ is a positive key strength value, produced by the head as well, used to attenuate or amplify the focus precision.

The location-based mechanism instead implements a rotational shift of the weighting. First, each head produces a scalar interpolation gate $g_t$ in the range $(0,1)$ used to blend between the old weighting $w_{t-1}$ and the weighting produced by the content-based addressing at this time step to produce the gated weighting $w_t^g$ as:

\begin{equation}
w_t^g=g_tw_t^c+(1-g_t)w_{t-1}.
\end{equation}

Then, the head produces a shift weighting $s_t$, defining a normalized distribution over the possible allowed integer shifts, for example using a softmax. This weighting is then combined with the gated weighting $w_t^g$ with a circular convolution:

\begin{equation}
\tilde{w}_t(i)=\sum_{j=1}^Nw_t^g(j)s_t((i-j)\Mod N).
\end{equation}

This operation can however cause the dispersion of the weighting over multiple locations, making them less focused, when the shifting is not sharp. To contrast this, an additional scalar $\gamma_t\geq1$ is produced by the head to sharpen the final weighting as:

\begin{equation}
w_t(i)=\frac{\tilde{w}_t(i)^{\gamma_t}}{\sum_{j=1}^N\tilde{w}_t(j)^{\gamma_t}}.
\end{equation}

The combination of these two addressing mechanism can result in three different behaviours:
\begin{enumerate}
\item The weighting from the content-based addressing can be used without any modification from the location-based mechanism,
\item The weighting from the content-based addressing can be shifted by the location-based one, allowing access to a specific element within a contiguous block of data,
\item The weighting from the content-based addressing is totally ignored, and the previous weighting is shifted by the location-based mechanism, resulting in an iteration over a sequence of addresses.
\end{enumerate} 

\section{Binary Arithmetic with NTMs}
\label{experiments}
The main goal of this work is to how well a standard NTM is able to learn and generalize two fundamental arithmetical tasks, the binary addition and the binary multiplication. The seminal work on NTMs \cite{ntm} shows how they are able to properly learn and generalize algorithmic tasks like the copy task or the associative recall working with a sequence of binary vectors as input, but these tasks do not explicitly require to combine the stored vectors in order to be successfully learnt, only to read them from the memory in the correct order. Arithmetical tasks, on the other hand, are suitable to assess if a NTM is capable of properly combing the values stored in its memory in order to get the correct result. This capability should in principle be ensured by "blurry" read and write operations, but for complex systems, theory and practise can diverge.

Binary arithmetical operations are a fundamental task in computer science, allowing computers to perform arbitrary operations by combining different pieces to compute simpler functions together. NTMs, with their ability to apply learned algorithms to sequences beyond the size of those used for training, poses themselves as a state-of-the-art mechanism to learn these tasks and generalize them.

\subsection{Experimental Settings}
For the described experiments, we represent binary numbers as sequences of bits in the little-endian notation, i.e. from the least significant bit to the most significant one. For example, the number $35$ is represented as $110001$ in the little-endian notation. We choose this notation, instead of the more common big-endian one, in order to feed the networks with the first bits they have to use in the output computation at the beginning.

At every time step, the input to the networks is a three-element vector encoded following the notation of Table \ref{tab:encoding}.

\begin{table}[htbp]
\centering
\begin{tabular}{|c|c|}
\hline
Symbol & Encoding \\
\hline
$0$ & $000$ \\
$1$ & $100$ \\
$+, *$ & $010$ \\
$END$ & $001$ \\
\hline
\end{tabular}
\caption{Binary encoding of an input vector.}
\label{tab:encoding}
\end{table}

Hence, an input sequence is formed as a binary number (i.e. a sequence of symbols $1$ and $0$), the symbol $+$ or $*$, another binary number and the $END$ symbol. The output sequence is formed by a binary number followed by the $END$ symbol. Moreover, because the tasks are episodic, we reset the memory matrix and the read vector of the NTMs as well as the hidden states of both the NTM controllers (when they are recurrent networks) and the LSTM layers after each training example. For example, the simple sum $8+3=11$ using $4$ bits is represented as:

\begin{equation*}
\underbrace{
\begin{pmatrix}
1 & 0 & 0 & 0\\
0 & 0 & 0 & 0\\
0 & 0 & 0 & 0\\
\end{pmatrix}}_{8}
\underbrace{
\begin{pmatrix}
0\\
1\\
0\\
\end{pmatrix}}_{+}
\underbrace{
\begin{pmatrix}
0 & 0 & 1 & 1\\
0 & 0 & 0 & 0\\
0 & 0 & 0 & 0\\
\end{pmatrix}}_{3}
\underbrace{
\begin{pmatrix}
0\\
0\\
1\\
\end{pmatrix}}_{END}
\underbrace{
\begin{pmatrix}
0 & 1 & 0 & 1 & 1\\
0 & 0 & 0 & 0 & 0\\
0 & 0 & 0 & 0 & 0\\
\end{pmatrix}}_{11}
\underbrace{
\begin{pmatrix}
0\\
0\\
1\\
\end{pmatrix}}_{END}
\end{equation*}

We compare two NTM architectures, using a feedforward network or an LSTM as the controller respectively with a standard LSTM network with $3$ hidden layers as a baseline. Table \ref{tab:parameters} shows a summary of the models details, taken from previous literature \cite{ntm,implementation}.

\begin{table}[htbp]
\centering
\begin{tabular}{|c|c|c|c|r|r|}
\hline
Architecture & \#Heads & Controller & Size & Memory Size & \#Parameters \\
\hline
FF-NTM1 & $1$ & Feedforward & $100$ & $128\times 20$ & $15,011$ \\
LSTM-NTM & $1$ & LSTM & $100$ & $128\times 20$ & $63,011$ \\
3h-LSTM & n.a. & n.a. & $3\times 128$ & n.a. & $333,059$ \\
\hline
\end{tabular}
\caption{Parameters of the considered architectures.}
\label{tab:parameters}
\end{table}

\subsection{Binary Addition}
The binary addition is a fundamental and one of the simplest arithmetical task. Algorithm \ref{algo:binary_add} shows how a human programmer would write an algorithm for the binary addition using little-endian notation.

\begin{algorithm}[htbp]
\footnotesize
\caption{Binary addition}
\KwData{two binary numbers $n1$ and $n2$ of length $l$}
\KwResult{their sum $s$ of length $l+1$}
$c=0$\;
\For{$i=1\ldots l$}{
$s[i]=(n1[i]+n2[i]+c)\mod{2}$\;
$c=(n1[i]+n2[i]+c)/{2}$\;
}
$s[l+1]=c$\;
\label{algo:binary_add}
\end{algorithm}

A model learning the binary addition task has to properly combine the considered bits of the two numbers as well as the carry from the previous step. It also has to compute both the new bit of the output and the (new) carry for the next step. This task, other than being a basic task in arithmetic and computer science, is not as simple as it may appear. Its complexity in time is $O(n)$. We can hypothesize that the internal algorithm learned by the NTMs, after that it had taken the whole two $n$-bits long numbers as input, should read from three memory locations (two for the numbers bits and one for the carry) and write into two locations (one for the new bit of the result and one for the carry) at every time step to complete the task in $n+1$ steps. For example, the binary addition $4+2=6$ using $3$ bits numbers is given by:

\begin{equation*}
\begin{bmatrix}
1 & 0 & 0
\end{bmatrix}+
\begin{bmatrix}
0 & 1 & 0
\end{bmatrix}=
\begin{bmatrix}
1 & 1 & 0
\end{bmatrix}
\end{equation*}
\\

\begin{figure}[htbp]
\centering
\includegraphics[width=0.65\textwidth]{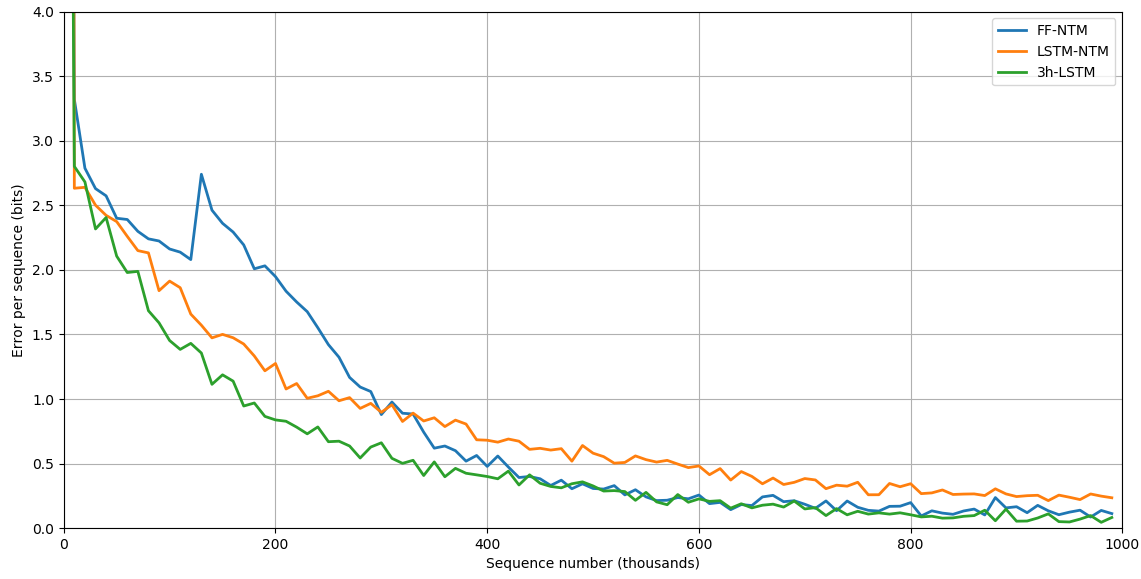}
\caption{Learning curves for the addition task.}
\label{fig:sum_training}
\end{figure}

The first thing we are interested in is to compare how well these models are able to learn the given task. We used the RMSProp algorithm \cite{rmsprop} with a learning rate $\alpha=10^{-4}$ and $\gamma=0.95$ as the training algorithm and the binary cross entropy as the objective function. We trained every model with $1,000,000$ examples, using binary numbers with a variable length up to $8$ bits. Figure \ref{fig:sum_training} shows the errors, expressed in bits per sequence, during the training process.

During the training process, FF-NTM achieves an overall smaller error (with some fluctuation at the beginning) with respect to both LSTM-NTM and the baseline. This could be due to the smaller number of parameters that could have speeded up the learning process. However, generally every proposed model performed well during the training, reaching an error of almost $0$ bits per sequence.

\begin{figure}[htbp]
\centering
\includegraphics[width=0.65\textwidth]{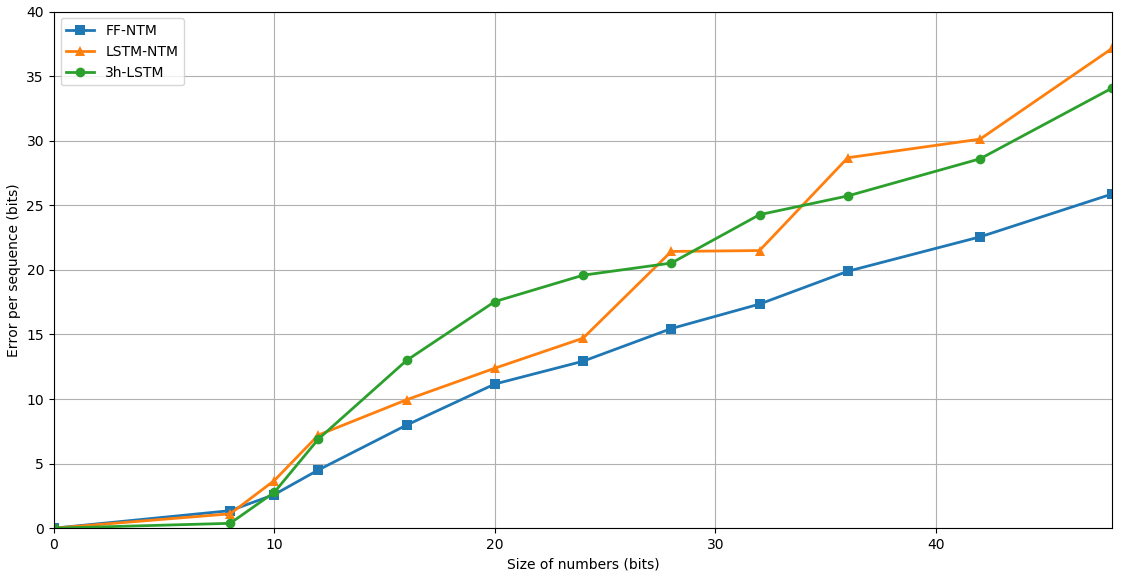}
\caption{Generalization error of the trained models.}
\label{fig:sum_errors}
\end{figure}

The second aspect we are interested in is to see how well the learnt algorithms generalize to larger input sequences than those used during the training process. We have trained the models with binary numbers of length up to $8$, so, considering also the $+$ and $END$ symbols, the sequences length vary from $4$ elements (when the two numbers are of length $1$) to $18$ (when the numbers are of length $8$). We have tested our models with numbers of length $8$, $10$, $12$, $16$, $20$, $24$, $28$, $32$, $36$, $42$ and $48$ (i.e. on sequences of length up to $98$ elements and with an output length of up to $n+1=49$ elements, therefore composed of $49\times 3=297$ bits). Figure \ref{fig:sum_errors} shows the mean error over $100$ tests.

From Figure \ref{fig:sum_errors} it is evident how FF-NTM largely outperforms the other models in terms of generalization ability, achieving a smaller reconstruction error on every sequence length. LSTM-NTM instead just performs slightly worse than the baseline, especially on longer sequences. This could be due to an insufficient training period or an inefficient use of the recurrent connections of the controller.

\begin{figure}[htbp]
\centering
\begin{minipage}{0.35\textwidth}
\centering
\includegraphics[width=0.8\linewidth]{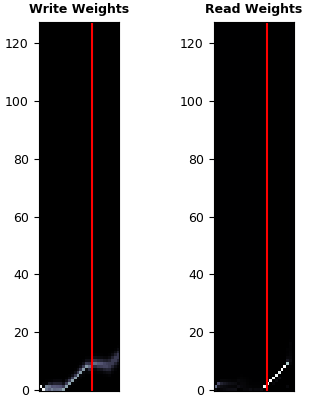}
\caption*{(a) FF-NTM}
\end{minipage}
\hspace{60pt}
\begin{minipage}{0.35\textwidth}
\centering
\includegraphics[width=0.8\linewidth]{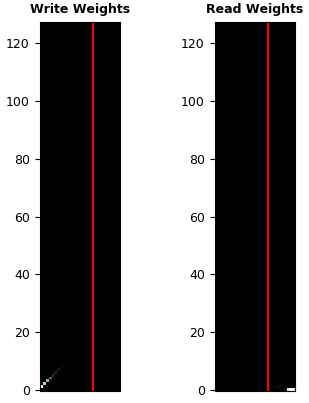}
\caption*{(b) LSTM-NTM}
\end{minipage}
\caption{Interaction with the memory for the FF-NTM (left) and LSTM-NTM (right) for the binary addition task.}
\label{fig:memory_sum}
\end{figure}

To try and understand what kind of algorithm the NTMs learned, we are now going to analyse their interactions with the memory for this task. In the following figure (and in the similar one obtained for the multiplication task) the time steps are intended from left to right in each of the two column. The white squares represent a fully focused interaction with a memory row at a certain time step, while black ones represent no interaction with that row. Grey squares represent interaction with a certain degree of "blurriness": the more it turns to white, the more focused the interaction with that memory row. The vertical red line highlights the moment in which the NTM has finished to receive the input. Figure \ref{fig:memory_sum} shows the reading and writing pattern for a $8$-bits long numbers sum.

FF-NTM uses a feedforward network as the controller, so it does not have recurrent connections to use as a sort of secondary memory: every time it needs to memorise something, it has to use the memory matrix. For this architecture, the pattern (on the left hand side of the figure) appears quite understandable. The network reads and stores the whole first binary number on $n$ consecutive rows and, when it starts to receive the second one, it contemporary reads the correspondent bit of the first one from the memory performs some computations, storing the the results back on the next rows of the memory.

LSTM-NTM instead, along with the memory matrix, can also store some information using its recurrent connections. This time the read and write pattern does not appear so clear: it start by memorizing something (presumably the first number) onto the memory, but it then gradually stops doing so, probably starting to rely on the recurrent connections. At the very end of the output sequence it just read one of the memory row. This highlight how different an algorithm learnt by and NTM can be with respect to the one we could think of as human programmer.

\subsection{Binary Multiplication}
The binary multiplication task is harder than the previously considered binary addition, because it requires to use various sums in order to get the final result. Algorithm \ref{algo:binary_mul} shows how a human programmer would probably write an algorithm for the binary multiplication.

\begin{algorithm}[htbp]
\footnotesize
\caption{Binary multiplication}
\KwData{two binary numbers $n1$ and $n2$ of length $l$}
\KwResult{their product $s$ of length $2*l$}
\For{$j=l\ldots l$}{
\If{$n1[j]=1$}{
$c=0$\;
\For{$i=1\ldots l$}{
$s[i]=(s[i]+n2[i]+c)\mod{2}$\;
$c=(s[i]+n2[i]+c)/{2}$\;
}
$s[l+1]=c$\;
}
}
\label{algo:binary_mul}
\end{algorithm}

So, in order to generalise the binary multiplication, a model has to learn how to sum up two numbers and when it has to perform the sum depending on the bits of one of the two numbers (see the {\tt if} statement in the code above). The cost of the algorithm is $O(n^2)$ in time, so it is more complex than the binary addition. For two $n$-bits long binary numbers, the length of the result is $2n$ bits.

\begin{figure}[htbp]
\centering
\includegraphics[width=0.65\textwidth]{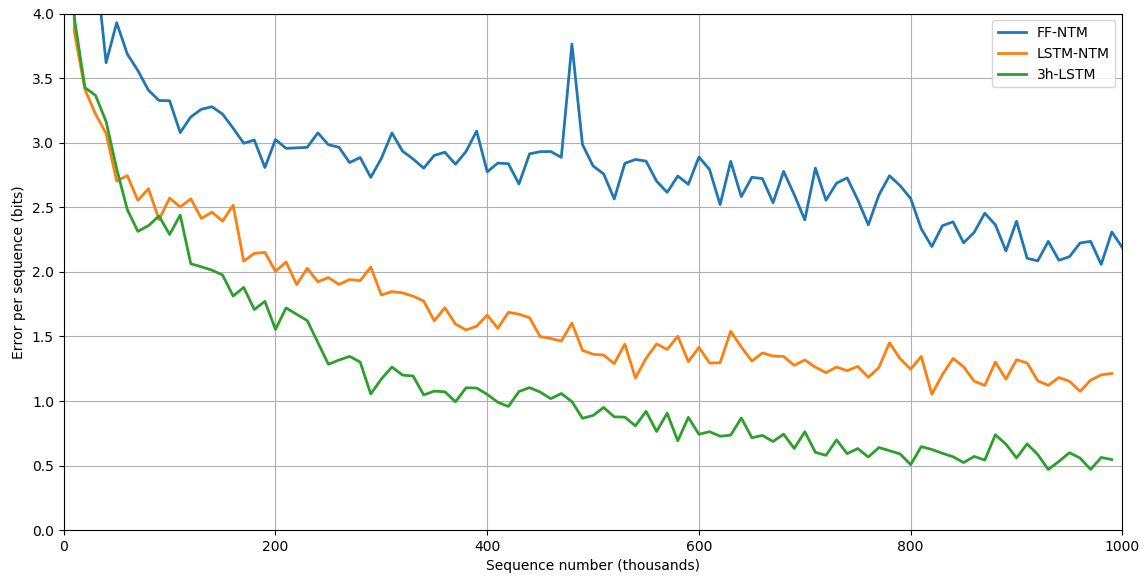}
\caption{Learning curves for the multiplication task.}
\label{fig:mul_training}
\end{figure}

For the training process, we again used the RMSProp algorithm with a learning rate $\alpha=10^{-4}$ and $\gamma=0.95$ as the training algorithm and we trained every model with $1,000,000$ examples, using binary numbers with a variable length up to $8$ bits as in the previous experiment. Figure \ref{fig:mul_training} shows the errors, expressed in bits per sequence, during the training process.

From the plot we can see how, during the training process, the baseline LSTM achieves a smaller error compared to the two proposed NTM models, especially compared to the FF-NTM one. This could be due to a simpler model, even with an higher number of parameters to be tuned, that can be trained easier for such a complex task.

\begin{figure}[htbp]
\centering
\includegraphics[width=0.65\textwidth]{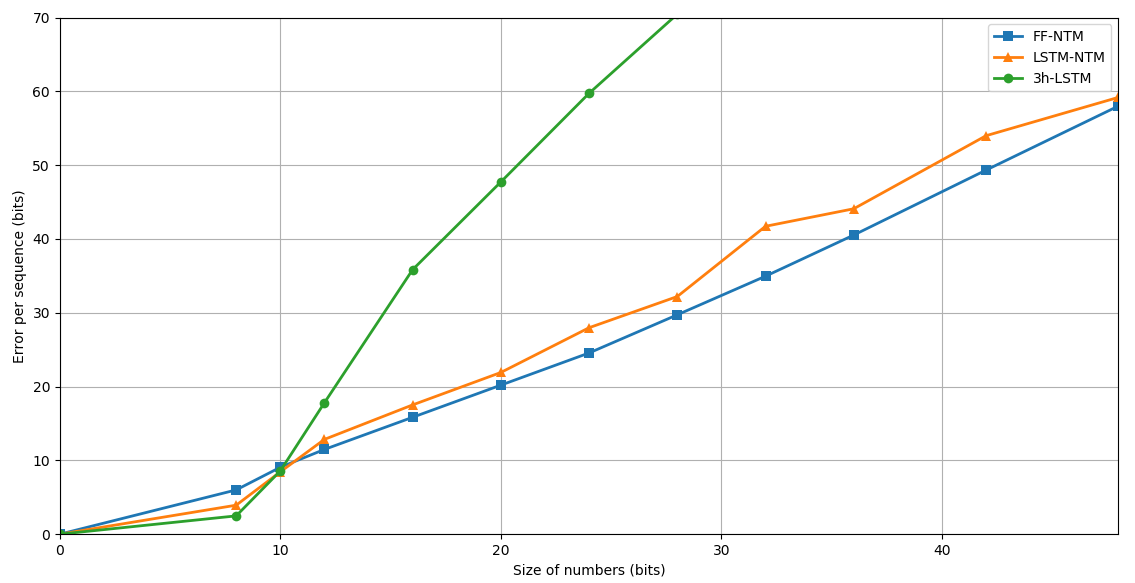}
\caption{Generalization error of the trained models.}
\label{fig:mul_errors}
\end{figure}

We have trained the models with binary numbers of variable length up to $8$ bits, so, considering also the $*$ and $END$ symbols, the sequences length vary from $4$ elements (when the two numbers are of length $1$) to $18$ (when the numbers are of length $8$). We have tested our models with numbers of length $8$, $10$, $12$, $16$, $20$, $24$, $28$, $32$, $36$, $42$ and $48$ (i.e. on sequences of length up to $98$ elements and with an output length of up to $2*n=96$ elements, therefore composed of $96\times 3=288$ bits), exactly as before. Figure \ref{fig:mul_errors} shows the the mean error over $100$ tests.

Figure \ref{fig:mul_errors} shows a huge gap in the generalization performance between the two NTM models and the baseline. In fact, while the formers behave in a similar way, performing just slightly worse than the baseline of shortest sequences and then maintaining a linear error increase over longer ones, the LSTM is not able to grasp the principles need to generalize the binary multiplication task, resulting is a very high error with sequences of just double the length of the ones used for training.

\begin{figure}[htbp]
\centering
\begin{minipage}{0.35\textwidth}
\centering
\includegraphics[width=0.8\linewidth]{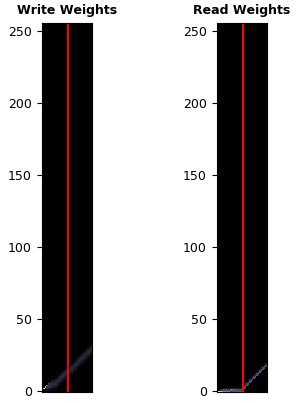}
\caption*{(a) FF-NTM}
\end{minipage}
\hspace{60pt}
\begin{minipage}{0.35\textwidth}
\centering
\includegraphics[width=0.8\linewidth]{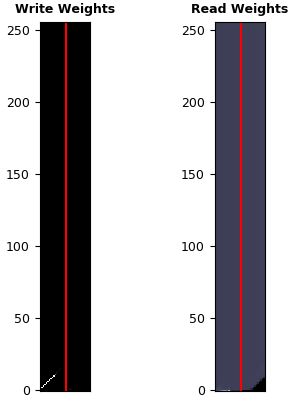}
\caption*{(b) LSTM-NTM}
\end{minipage}
\caption{Interaction with the memory for the FF-NTM (left) and LSTM-NTM (right) for the binary multiplication task.}
\label{fig:memory_mul}
\end{figure}

We now try to analyse the interactions of the tested models with their memory for the binary multiplication task. Figure \ref{fig:memory_mul} shows the reading and writing pattern of the various NTM models for the product of two $8$-bits long binary numbers.

Also for this task we face a situation similar to the one we had for the binary addition task. FF-NTM learns a sparse read and write pattern that could mean that the controller is really mixing the two numbers while getting them as input, storing the various intermediate results needed for the computation. The LSTM-NTM model instead again produces a not clear pattern, starting by storing into the memory the first number it receives, but then probably relying on its recurrent connections in order to complete the computation. The reading pattern in particular is not clear: the controller seems to sparsely read over all the memory matrix, without focusing on any location (and therefore any stored information) specifically. This further support the idea that the controller is mainly using its recurrent connections rather than the external memory.

\section{Conclusions}
\label{conclusions}
Neural Turing Machines are a novel approach to machine learning first designed by Google DeepMind Labs in 2014 \cite{ntm}. By coupling a standard neural network with an external portion of memory, they tried to overcome some of the difficulties found by standard RNNs to remember and exploit long time recurrences. 

The primary objective of this work was to test and understand how well a NTM can learn two fundamental arithmetical tasks that had not been previously investigated: the binary addition and the binary multiplication. These two problems are harder than the tasks previously learned with this approach, so they represent a challenge to this model other than being interesting on their own. We think that we can give a positive answer to this question: some of the proposed architectures really outperform standard LSTMs in terms of learning capability and generalization, beyond the fact that they have less parameters to be tuned than the considered LSTM baseline itself. Unfortunately, the algorithms they have learned to archive these results are not always simple to understand, highlighting the difficulties in predicting the behaviour of neural networks as well as the main differences in designing an algorithm between a human and the latter.

As a future line of work, it would be interesting to try and gain a deeper understanding of the algorithms learnt by these models, as well as coupling the proposed standard architecture with some of the extensions proposed in literature \cite{lie,memory,attention,structured} or with an higher number of read and write heads, to assess if they can help to improve the generalization performances.

\nocite{*}
\bibliographystyle{plainnat}
\bibliography{Bibliography}
\end{document}